\documentclass[Afour,sageh,times]{sagej}

\usepackage{moreverb,url}

\usepackage[colorlinks,bookmarksopen,bookmarksnumbered,citecolor=red,urlcolor=red]{hyperref}

\usepackage{amsmath,amsfonts}
\usepackage{subcaption}
\newcommand\BibTeX{{\rmfamily B\kern-.05em \textsc{i\kern-.025em b}\kern-.08em
T\kern-.1667em\lower.7ex\hbox{E}\kern-.125emX}}

\def\volumeyear{2016}
\newtheorem*{definition*}{Definition:}

\begin{document}

\runninghead{Bhagat Smith and Adams}

% \title{Workload Estimation for Unknown Tasks: A Survey of Machine Learning Under Distribution Shift}
\title{A Survey of Machine Learning for Estimating Workload: Considering Unknown Tasks}

\author{Joshua Bhagat Smith\affilnum{1} and Julie A. Adams\affilnum{1}}

\affiliation{\affilnum{1}Oregon State University, OR, USA}

\corrauth{Julie A. Adams, Oregon State University
Collaborative Robotics and Intelligent Systems (CoRIS) Institute,
Corvallis, OR, USA.}

\email{julie.a.adams@oregonstate.edu}

%Reframe this as an arguement for how to use these techniques
\begin{abstract}
% Old abstract
% Human-robot teams involve humans and robots collaborating to achieve tasks under various environmental conditions. Successful teaming will require robots to adapt autonomously to a human teammate's internal state. An important element of such adaptation is the ability to estimate the human teammates' workload in unknown situations. Existing workload models use machine learning to model the relationships between physiological metrics and workload; however, these methods are susceptible to individual differences and are heavily influenced by other factors. These methods cannot generalize to unknown tasks, as they rely on standard machine learning approaches that assume data consists of independent and identically distributed (IID) samples. This assumption does not necessarily hold when estimating workload for new tasks. A survey of non-IID machine learning techniques is presented, where common techniques are evaluated using three criteria: portability, model complexity, and adaptability. These criteria are used to argue which techniques are most applicable for estimating workload for unknown tasks in dynamic environments.

Successful human-robot teaming will require robots to adapt autonomously to a human teammate's internal state, where a critical element of such adaptation is the ability to estimate the human's workload in unknown situations. Existing workload models use machine learning to model the relationship between physiological signals and workload. These methods often struggle to generalize to unknown tasks, as the relative importance of various physiological signals change significantly between tasks. Many of these changes constitute a meaningful shift in the data's distribution, which violates a core assumption made by the underlying machine learning approach. A survey of machine learning techniques designed to overcome these challenges is presented, where common techniques are evaluated using three criteria: portability, model complexity, and adaptability. These criteria are used to analyze each technique's applicability to estimating workload during unknown tasks in dynamic environments and guide future empirical experimentation.
\end{abstract}


\maketitle

% 1. Reframe the message
\section{Introduction}

Deploying human-robot teams in uncertain environments requires robots to have a dynamic understanding of humans' internal state. This dynamic understanding must account for real-world complexities in order to enable fluid interactions between a robot and a human teammate. Workload represents how hard a person is working, and can be decomposed into workload components (i.e., cognitive, speech, auditory, visual, gross motor, fine motor, and tactile). 

Incorporating machine learning-based workload models that accurately capture the contribution of each component will enhance the robot’s understanding of its human teammate, allowing it to respond appropriately to undesirable workload levels (i.e., overload (OL) or underload (UL)) through intelligent modulation of how the robot interacts with the human. These adaptive interactions enable more collaborative human-robot teaming dynamics and facilitate long-term collaboration. However, real-world human-robot teams will inevitably encounter novel (i.e., unknown) tasks. These tasks present unique challenges for the machine learning models that underpin modern workload estimation algorithms that rely on physiological signals, as unknown tasks frequently exhibit unique data characteristics that the model was not trained to account for.

An adaptive teaming system requires machine learning models that can estimate workload accurately in both known and unknown situations. Further, workload estimates are susceptible to individual differences across individuals performing the same task, the same individual performing different tasks, and even for the same individual performing the same task on different days \citep{longo2022human, wickens2004introduction}. Workload is also influenced by factors, such as experience, stress, and fatigue \citep{heard2019diagnostic}. Developing machine learning-based workload estimation models that generalized to novel situations while also accounting for these factors is difficult, as these traditional approaches assume data consists of independent and identically distributed (IID) samples \citep[chap. 19]{pml2Book}. The IID assumption states that a machine learning model's training and testing data (e.g., real-world data) are drawn from the same probability distribution. This assumption implies that the training data is sufficiently representative of real-world data, which does not necessarily hold when estimating workload for novel, unknown tasks.

Prior work established that workload models built using standard machine learning methods are capable of generalizing across individuals for similar tasks \citep[e.g.,][]{heard2019diagnostic, kaczorowska2021interpretable, manawadu2018multiclass}, but there is no widely accepted approach that can generalize to novel, unknown tasks \citep{longo2022human}. Generalizing across tasks is difficult due to a shift in the relative balance of workload components, where certain physiological signals and workload components are of particular importance to a given task and the magnitude of that significance varies between tasks \citep{heard2019diagnostic}. Consider first response robotics. Urban firefighters will work alongside robots to perform many tasks \citep{delmerico2019current}. Firefighters have historically used robots to gather aerial imagery, but advanced control algorithms will enable robots to extinguish fires in otherwise  inaccessible locations \citep{perez2023urban}. This new task will require different cognitive and visual demands, as well as different degrees of robot control, supervision, and interactions. 

The dynamic relationship between physiological signals and an individual's underlying workload makes the development of machine learning models that generalize in the real world difficult \citep{albuquerque2022estimating, popivanov1999testing}. Prior work demonstrated that this dynamism is considerably influenced by contextual factors (e.g., tasks demands, teaming dynamics), as contextual factors heavily influence an individual's workload \citep{heard2018survey, longo2022human} and their physiological response to external stimuli \citep{larradet2020toward, saganowski2022emotion}. Considering how contextual factors influence each workload components adds additional complexity. Each component relies on a unique subset of physiological signals \citep{bs2022physical, bs2022visual, fortune2020real, heard2019diagnostic}, but it is impossible to predict the specific physiological response novel tasks will incur. Gathering the necessary data to develop workload estimation models exacerbates this problem, as experimental human-subject evaluations rarely capture the fully breadth of real-world human activity \citep{bs2024design}.

The variability between contexts (e.g., tasks) violates the IID assumption \citep{albuquerque2022estimating, cao2014non, popivanov1999testing}, and must be analyzed through the lens of \emph{distribution shifts} (i.e., the mathematical differences between datasets). Prior work developed a broad range of non-IID machine learning techniques (e.g., continual learning, meta-learning) that incorporate contextual examples to ameliorate a distribution shift's negative impact \citep{quinonero2022dataset}. These techniques have demonstrated success in many application domains (e.g., computer vision \citep{tian2020rethinking}, natural language \citep{latif2022self}, task recognition \citep{leite2022resource}), but their success varies based upon several dataset characteristics (e.g., labeling, volume). There exists too many techniques for a comprehensive evaluation to be tractable; thus, a more careful analysis evaluating how relevant human factors interact with a technique's machine learning characteristics is required. 

The manuscript surveys applicable non-IID machine learning methods. These methods are evaluated to assess their viability for developing workload estimation models that can accommodate real-world variability. First, the fundamentals of objective workload estimation and distribution shifts are presented, followed by requisite notation and criteria. These criteria are used to analyze a wide range of non-IID machine learning techniques, where arguments for which techniques merit empirical investigation are presented. The most applicable techniques are further discussed, which highlights the importance of jointly evaluating a problem's machine learning and human factors considerations in order to determine a techniques real-world applicability. Finally, conclusions and future work are presented.

% 2. Shift the references away from ML to HF, in order to drive that message home.
\section{Background}

Workload is a complex, dynamic, individual-specific, non-linear construct, and developing models that can account for these complexities is a non-trivial task. Fundamentally, estimating workload for unknown tasks is a problem of generalization and recent work has established that existing workload assessment methods encounter difficulties generalizing across tasks, either known or unknown \citep{longo2022human, zhao2018real}. 

% Modern techniques (e.g., meta-learning) approach machine learning from a different perspective and possess the potential to build more flexible workload estimation models.

Humans have limited resources for processing information, making decisions, and dealing with physical stress. Formally, workload is the degree of activation of a finite pool of resources while performing a task \citep{longo2022human} that is influenced by environmental factors, situational factors, and internal characteristics (e.g., fatigue, experience). Workload has been shown to vary across individuals, across tasks, and over time \citep{christensen2012effects, wickens2004introduction}. Workload can be divided into different components based upon the types of resources being utilized: \emph{cognitive, speech, auditory, visual, and physical} \citep{mitchell2000mental}, and physical can be further subdivided based upon the nature of physical work: \emph{gross motor, fine motor, and tactile} \citep{heard2018survey}.

Robots operating in dynamic environments need high-frequency workload measures to adequately adjust to their human teammates' current state. Workload estimation methods either rely on subjective or objective metrics. Surveys and questionnaires (e.g., NASA Task Load Index \citep{hart1988development}, In-Situ Surveys \citep{wilson2003real}) are common subjective metrics that rely on an individual to self-assess their perceived workload levels \citep{heard2018survey}. Subjective metrics fail to provide a continuous measure and are sensitive to the bias inherent to self-reports \citep{kosch2023survey, matthews2020subjective}; thus, they are not viable for real-world environments. Physiological signals provide a quantitative and consistent method for measuring a human's workload and are the most widely used objective metric \citep{debie2019multimodal, guan2022eeg}. Objective workload estimation methods use machine learning to learn the relationship between physiological metrics, collected via wearable sensors, and an individual's underlying workload \citep[e.g.,][]{cao2021recognition, guo2021eye, quan2023eeg}. These methods provide workload estimates at the necessary frequency and accuracy for real-world environments \citep{bagheri2022simultaneous, fortune2020real, heard2019diagnostic}. 

Many objective workload estimation methods are ill suited for uncertain, dynamic environments as the sensors required to collect the necessary physiological signals are either complex systems (e.g., electroencephalogram (EEG) \citep{ved2021detecting}) or environmentally embedded (e.g., built-in cameras \citep{kosmopoulos2012system}). The human must be able to move through the environment freely. Environmentally embedded sensors restrict the human's movement and complex sensor systems are extremely sensitive to the noise introduced by human movement, making them ill suited for uncertain, dynamic environments. Wearable sensors (e.g., a heart-rate monitor) can serve as the primary data source, enabling the human and the robot to act independently. Wearable sensors are susceptible to high sensor noise and latency issues, but prior work has demonstrated their ability in accurately estimating workload in real-time environments \citep{dayal2024novel, heard2020sahrta, liu2024classifying, park2024pilot}.

Robots need a reliable estimate of the distribution of workload across the workload components, as tasks will change over time, and differing tasks impact different components. Each workload component corresponds to a different set of physiological metrics, though some metrics may correspond to multiple components. Incorporating multiple wearable sensors and physiological signals allows a workload estimation method to differentiate between the workload components and accurately measure their contribution to overall workload \citep{heard2018survey, bs2024design}.

Prior work established four metrics for evaluating workload estimation algorithms: Sensitivity, Diagnosticity, Suitability, and Generalizability \citep{heard2018survey}. Sensitivity represents an algorithm's ability to reliably detect workload levels (i.e., \(\geq\) 80\% accuracy or \(\leq\) 5 root mean squared error (RMSE)). Diagnosticity refers to the capability of detecting different types of workload. Suitability refers to an algorithm's ability to assess the complete overall workload state. The last metric, generalizability, refers to the algorithm's ability to generalize across individuals and tasks. This metric defines ``generalizing across tasks" as an algorithm's capacity to asses all seven workload components and defines ``generalizing across individuals" as an algorithm's capacity to maintain an accuracy of \(\geq\) 80\% or an RMSE \(\leq\)5 for all individuals.

% It is important to note the nuances between diagnosticity and suitability. Consider a driving task that incurs a significant cognitive and visual workload. An algorithm that produces a measurement of overall workload is considered suitable, but not diagnostic. An algorithm is only considered diagnostic if it is able to enumerate the specific contributions of the cognitive and visual components to overall workload. 

Most existing methods fail to meet these criteria, as they primarily perform binary classification of cognitive workload \citep[e.g.,][]{dell2021mbiotracker, hogervorst2014combining, xie2019personalized, yin2019physiological}. Many methods are either exhibit suitability \citep[e.g.,][]{islam2020novel, momeni2019real, moustafa2017assessment, zhang2020instantaneous} or sensitivity \citep[e.g.,][]{caywood2017gaussian, ding2020measurement, kaczorowska2021interpretable, ved2021detecting}, but very few exhibit diagnosticity. Further, little work evaluated the generalizability of existing workload estimation methods.

% An algorithm needs to generalize across tasks, while only relying on wearable sensors, to effectively estimate workload in dynamic environments. There are several methods for estimating workload using wearable sensors (e.g., \citep{heard2019diagnostic, kaczorowska2021interpretable, manawadu2018multiclass}). The generalizability criteria defines generalizing across tasks as an algorithm's capacity to estimate the complete workload state \citep{heard2018survey}. Heard et al. developed a diagnostic workload algorithm that achieved high performance levels on known tasks \citep{heard2019diagnostic}; however, this algorithm does not estimate the complete workload state, as it heavily relies on static models for visual workload. Thus, it does not strictly meet the generalizability criteria. Further, the generalizability criteria's definition of generalizing across tasks is too narrow, as it ignores the task's context. An algorithm's ability to generalize across tasks must also consider the accuracy, or RMSE of workload estimates for known and unknown tasks. 

Prior work tends to focus on generalizing across individuals. Many algorithms employed a leave-one-subject-out cross-validation strategy to validate generalization across individuals \citep[e.g.,][]{albuquerque2019cross, guo2021eye, heard2019diagnostic, hefron2017deep, novak2015workload}; however, generalizing across tasks, either known or unknown, is rarely evaluated. Algorithms specifically developed to generalize across tasks have received increased attention, and are typically referred to as \emph{cross-task workload estimation} algorithms \citep{appel2021cross}. Many of these efforts evaluated an existing algorithm's ability to generalize across tasks, but those algorithms did not achieve either a \(\geq 80\%\) accuracy, or a \(\leq\) 5 RMSE on arbitrary tasks \citep[e.g.,][]{appel2021cross, baldwin2012adaptive, besson2012bayesian, boring2020continuous, walter2013using}. Thus, these algorithms exhibit sensitivity, but not generalizability. Other algorithms achieved \(\geq 80\%\) accuracy across tasks, but only perform binary classification \citep[e.g.,][]{dimitrakopoulos2017task, kakkos2021eeg, zhang2018learning}; thus, failing to meet the generalizability criterion.

Some support vector machine-based workload estimation algorithms have achieved achieved \(<\) 1.0 mean squared error on cross-task workload estimation \citep{guan2022eeg, ke2015towards, zhou2022cross}. Deep learning techniques (e.g., attention mechanisms) that estimate workload have also shown promise \citep{taori2022cross}. However, the majority of existing methods focus solely on cognitive workload and rely on high data volumes from complex or environmentally embedded sensors \citep[e.g.,][]{ji2023cross, guan2023cross, wen2023cross}. Thus, none of the methods are viable for use in dynamic, uncertain environments.

Real-world human-robot teams will inevitably encounter unique tasks when deployed. An adaptive teaming system requires models that can accurately estimate workload in both known and unknown scenarios, as well as accommodate individual differences \citep{wickens2004introduction}. Generalizing to unknown tasks while also accounting for individual differences is difficult for traditional machine learning, as the standard approaches assume data consists of independent and identically distributed (IID) samples \citep[chap. 19]{pml2Book}. Generalizing across tasks is difficult primarily due to a shift in the relative balance of the workload components associated with a task; thus, various physiological signals are of particular importance to a specific task and the magnitude of that significance varies between situations \citep{heard2019diagnostic}. This shift in the balance of workload components and physiological signals violates the IID assumption, which requires methods that use non-standard machine learning techniques \citep[chap. 19]{pml2Book}.

\section{Notation and Evaluation Criteria}

Dynamic human-robot teams will employ robots that estimate their human teammate's workload using machine learning methods. Extending the robot's capabilities to accommodate unknown tasks poses unique machine learning challenges, as these tasks constitute a meaningful distribution shift. A \emph{distribution shift} indicates a scenario where training and testing data come from different probability distributions. Understanding the difference between applicable machine learning methods is key for building human-robot teams that can handle the real-world complexity. The workload estimation and the machine learning literature both cover a large spectrum of work. There are frequently occurring terms in both fields; thus, analyzing applicable methods requires disambiguating key terminology and establishing evaluation criteria to ensure consistent analysis.

\subsection{Notation}

The term \emph{task} in the workload literature refers to a job an individual may perform (e.g., lifting boxes), but the machine learning literature uses the term to refer to a learning task for a machine learning model. The term \emph{domain} is used to characterize the input space for a machine learning problem. For example, performing object detection under various weather conditions (e.g., sunny, cloudy) can be viewed as two domains characterized by the weather \citep{sun2022shift}. Domain can also refer to a set of activities a human performs for a particular job (e.g., wildland firefighters). 

The term \emph{learning task} will refer to a machine learning problem. The term \emph{task}, without adjectives, will refer to an activity conducted by a human. The term \emph{domain} will refer to different input spaces that a machine learning algorithm may operate on. The term \emph{application domain} will refer to applications where human-robot teams may be deployed. 

The definitions of domain and learning task, in the context of distribution shift, are established \citep{pan2009survey}:

%\vspace{2.5pt}

\begin{definition*}\textbf{(Domain)} A \emph{domain} \(D\) is a two-tuple \((\mathcal{X}, P(X))\). \(\mathcal{X}\) is the feature space of \(D\) and \(P(X)\) is the marginal distribution, where \(X = \{x_1, \dots, x_n\} \in \mathcal{X}\).
\end{definition*}

%\vspace{2.5pt}

\begin{definition*}\textbf{(Learning Task)} A \emph{learning task} \(L\) is a two-tuple \((\mathcal{Y}, f(x))\) associated with a specific domain \(D\). \(\mathcal{Y}\) is the label space of \(D\) and \(f(x)\) is an objective predictive function for \(D\), which is learned from the training data. \(f(x)\) can be written as a conditional probability distribution \(P(y|x)\).
\end{definition*}

%\vspace{2.5pt}

Grounding this notation in workload estimation provides an explanation of the different variables. A particular workload estimation instance represents the domain, \(D\), which is characterized by the set of physiological metrics, \(\mathcal{X}\). The marginal distribution, \(P(X)\), represents the probability distribution over the physiological metrics' possible values. 

A learning task, \(L\), is a machine learning problem that is characterized by a label space \(\mathcal{Y}\) and a predictive function \(f(x)\), corresponding to a single domain \(D\). The label space represents a set of possible workload values, and the predictive function produces a workload value, \(y_i \in \mathcal{Y}\) given a set of physiological metrics, \(x_i \in \mathcal{X}\). A predictive function can take the form of a conditional probability distribution \(P(\mathcal{Y}|\mathcal{X})\), which represents the probability of assigning a workload value to a given set of physiological metrics. 

It is important to note that changes to either \(D\), \(\mathcal{Y}\), or \(f(x)\) constitute a new learning tasks. Changes to a domain (i.e., \emph{domain shift}) are intuitive, as they represent a distribution shift of physiological metrics brought on by performing a task under different working conditions (e.g., weather). Changes to a label space (i.e., \emph{label shift}) are characterized by a different range of possible workload values or distribution across workload components, typically brought on by performing a new task. Changes to the predictive function \(f(x)\) are less intuitive and more easily explained via task recognition. 

A typical task recognition problem is defined as training a predictive function to learn the relationship between a set of physiological metrics and a label space (e.g., [1, 2, and 3]), where each integer represents a unique task. A similar task recognition problem can be use the same metrics and the same label space, but with each integer representing the task difficulty (i.e., [easy, medium, or hard]). These two problems are semantically different, but the underlying machine learning method does not inherently recognize these differences; therefore, the uniqueness of these two learning tasks cannot be fully specified by the domain and label space. The changes in how the resulting predictive functions, \(f(x)\), map the physiological metrics to the output values is a key characteristic in differentiating these two learning tasks.

\subsection{Evaluation Criteria} \label{sec:criteria}

\begin{figure*}[!htp]
    \centering
    \includegraphics[height=7.25cm]{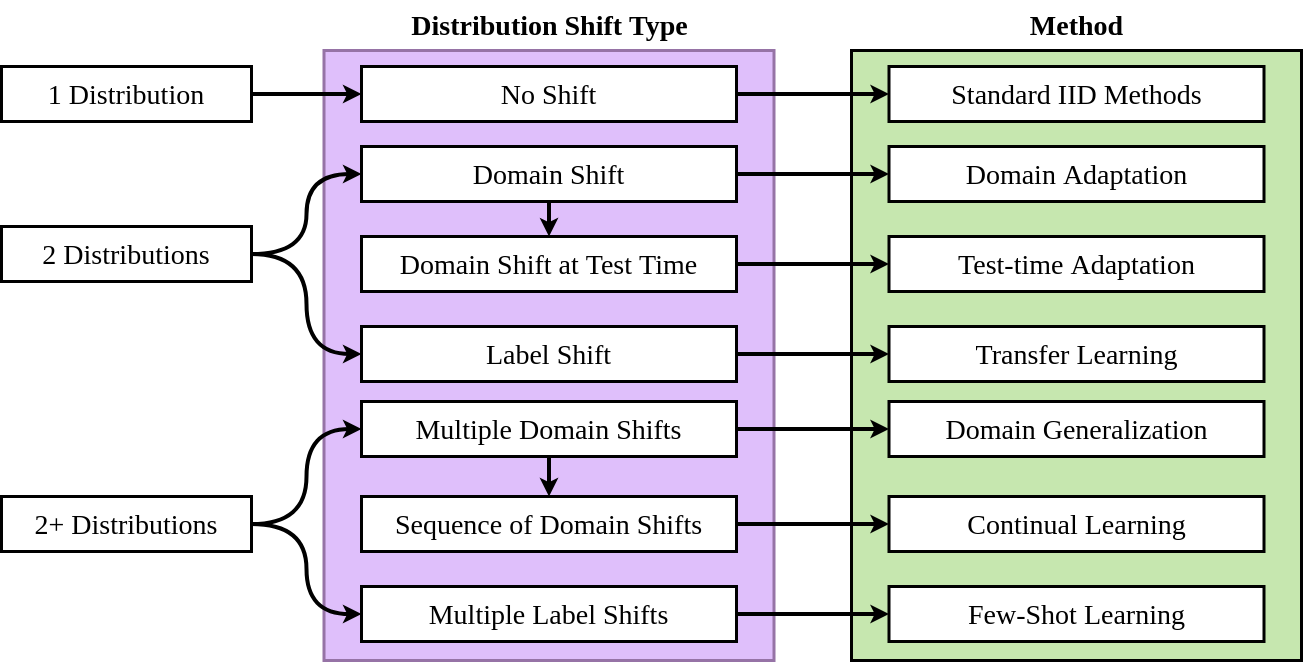}
    \caption{A high level overview of non-IID machine learning methods. Adapted from Figure 19.11 of \citep[chap. 19]{pml2Book}.}
    \label{fig:dist_shift}
\end{figure*}

An exhaustive analysis of applicable machine learning methods is impractical due to the volume of existing methods applied to non-IID machine learning problems; however, these methods can be conceptually grouped based upon the number of distributions and the distribution shift type (see Fig. \ref{fig:dist_shift}) \citep{pml2Book}. These categories represent a comprehensive list of non-IID machine learning method types and each category is assessed to understand which method types merit further investigation. Three evaluation criteria are applied to the presented techniques: (1) Portability, (2) Model Complexity, and (3) Adaptability. These three criteria function as a tool for understanding how the human factors considerations of workload estimation for unknown tasks interact with the machine learning aspects of non-IID techniques, with the aim of determining each technique's practical utility.

\emph{Portability} refers to the number of distributions used to train and test a model \citep{sun2015survey}. Learning tasks may consist of multiple source domains (e.g., urban vs wildland fire fighting). A straightforward means of combining information from multiple domains is to construct a single joint distribution; however, data in some domains may be of higher quality than others. Weighting high-quality data over low-quality data is appropriate to maximize performance \citep{sun2015survey}. Additionally, the separation of source domains may be artificial. Prior work established that modeling domains separately can be beneficial for learning inductive biases that improve performance on unknown learning tasks \citep{baxter2000model}. Portability can have four values: one-to-one, one-to-many, many-to-one, and many-to-many. \emph{One-to-one} refers to an algorithm that transfers knowledge between one source and one target domain. \emph{One-to-many} transfers knowledge from one source domain to many target domains, and \emph{many-to-one} transfers knowledge from multiple source domains to a single target domain. \emph{Many-to-many} knowledge transfer occurs between multiple sources to multiple target domains simultaneously.

A many-to-many portability is desirable for workload estimation for unknown tasks. Machine learning-based workload estimation methods are primarily developed using data collected in experimental conditions, producing diverse datasets with low data volumes. Further, these experimental conditions may unintentionally introduce bias into the dataset \citep{bs2024design}. Developing workload estimation methods must therefore be able to accommodate distribution shifts for multiple, small, noisy datasets (i.e., many-to-many portability). Many-to-one portability is only desirable if a method needs to estimate workload for a single new task, which is unlikely in real-world application domains. One-to-one portability is not sufficient outside of contrived circumstances, as effectively capturing the nuanced differences between known and unknown tasks using two distributions requires large data volumes.

\emph{Model Complexity} refers to the number of parameters an underlying machine learning model needs for the algorithm to be successful \citep{goodfellow2016}. Machine learning models with more parameters require higher volumes of training data. Additionally, training machine learning models in human-robot interaction (HRI) application domains requires this data to be ecologically valid, introducing additional complexity to model development. Early statisticians built simple linear models using hundreds of examples \citep{goodfellow2016}, whereas modern deep learning architectures may require tens of billions of examples \citep{brown2020language}. Model complexity shares a relationship with both the underlying machine learning model type and problem complexity, as complex models are often required to solve complex problems. Workload estimation and many HRI problems are typically conducted in low-data regimes, as they require human subject evaluations for data collection; thus, smaller models are preferred. An example of low model complexity is a random forest model that contains a few hundred decision trees and typically contain on the order of thousands of parameters \citep{breiman2001random}. An example of medium model complexity is shallow neural networks (i.e., fully connected neural networks) that typically contain hundreds of thousands of parameters \citep{fortune2020real}, but may contain on the order of millions of parameters. Two examples of high model complexity are deep learning and generative AI architectures (e.g., convolutional networks, long short-term memory networks (LSTM), transformers). These architectures can contain anywhere from tens of millions \citep{wilson2020survey} to over 100 billion parameters \citep{brown2020language}.

A machine learning-based workload estimation method must be capable of running onboard a robot to be useful in uncertain, dynamic environments. The robot is not guaranteed to have a remote connection to a powerful computer or access to the cloud. Modern edge compute solutions (e.g., NVIDIA Jetson Orin \citep{nvidia}) have the capability of executing many deep learning architectures in real-time, but it is desirable to minimize the model complexity in order to reduce the requisite computational resources and eliminate reliance on cloud resources. Further, workload estimation methods will be incorporated into a larger human-robot teaming architecture that may rely on other machine learning models. Reducing the model complexity of the workload estimation method frees up resources required by other components. Minimizing model complexity reduces the power consumed by the machine learning methods allow for more efficient power consumption, a critical constraint in uncertain, dynamic environments. Smaller models also require fewer computational resources that may be necessary for other learning tasks (e.g., task recognition, decision making).

The last metric, \emph{Adaptability}, refers to the amount of data required to update a machine learning model for new learning tasks \citep{wang2020generalizing}. Many non-IID methods seek to account for distribution shifts by updating existing machine learning models' parameters. Adaptability is the number of test data points necessary for the updated model to make accurate predictions for new learning tasks, where the goal is to minimize the required data. A machine learning method has no adaptability if it requires the same amount of training data in the source and target domains. Prior work suggested that 5000 examples per class is a sufficient rule of thumb for training deep learning architectures from scratch to achieve a minimum acceptable performance \citep{goodfellow2016}. A 50\% reduction in data required for updating a model is non-trivial, and may still constitute a large dataset. Therefore, an algorithm is considered to have low adaptability if it requires \(>\) 2500 examples per class (i.e., many) to optimize a function in a target domain \citep{zhuang2020comprehensive}.  An algorithm has high adaptability if it requires \(<\) 10 (i.e., a few) examples to update a model, a small dataset in all contexts \citep{wang2020generalizing}.

Adaptability is application-domain specific, as learning tasks are fundamentally different. Real-world human-robot teams will encounter unknown tasks in the field and must be able to estimate the human's workload as quickly as possible. Achieving a high adaptability is ideal, requiring a workload estimation method to collect large amounts of data in dynamic environments is impractical.

% 3. Ground discussion in a few good examples. Always refer back to the relevance of HF or HRI.
\section{Non-IID Machine Learning Methods}

% "How were these selected?" So we need citations and mention that while these broad categories are comprehensive, but each categories is a field of research unto itself. This paper seeks to understand the high-level characteristics of each category and argue for the most relevant ones to motivate future inquiry. 

Non-IID machine learning methods seek to account for distribution shifts by intelligently incorporating information from multiple distributions (i.e., domains) and fall into six categories: domain adaptation, transfer learning, test-time training, continual learning, domain generalization, and few-shot learning (see Figure \ref{fig:dist_shift}). These categories are primarily differentiated by their data handling procedure, which can also be viewed as how each method formulates the distribution shift problem \citep[chap. 19]{pml2Book}. Analyzing each category's data handling procedures is central to understanding a method's practical utility for estimating workload associated with unknown tasks, as it provides insight into how these data handling procedures can take advantage of the unique characteristics of HRI datasets.

Datasets used to train machine learning-based workload estimation methods are heavily impacted by the variability inherent in the physiological signals, noise within wearable sensor data, and range of potential human behaviors \citep{das2024cognitive}. These data characteristics present challenges unique to HRI application domains and differentiate them from other learning tasks (e.g., object detection, speech recognition). Additionally, most machine learning-based workload estimation methods rely on datasets collected in controlled experimental conditions. 

Methods within each category implement similar data handling procedures, where each category varies based upon the distribution shift type and number of distributions considered. There are two primary distribution shift types. A domain shift indicates a distribution shift of the input features (e.g., heart-rate variability, pupil diameter), and label shift (e.g., workload value) indicates a shift of the output values. Domain shifts can be characterized by changes to the range of values in various physiological signals or changes to the relative importance for any given physiological signal for estimating workload. Label shifts can be characterized by either a new distribution of output values, or a new set of values. For example, consider a training dataset where the cognitive workload values range from 0 to 15, with a mean value of 10 and a median value of 5. A testing dataset with a label shift may have cognitive workload values ranging from 5 to 20, or may have a mean and median value of 14. The degree to which these distribution shifts impact performance varies based upon the shift's magnitude and dataset size.

Non-IID machine learning methods either account for exactly two distributions or more than two distributions, where the boundaries between distributions may be real or artificial. Human behavior is most aptly described by learning tasks with multiple distribution shifts across domains, as humans perform diverse tasks in different ways in a wide range of environments. However, the data can be decomposed or re-organized such that any of the presented techniques can be applied. Data from a single, noisy distribution may be decomposed into multiple distributions to improve generalization to new tasks \citep{vinyals2016matching}, or data from multiple distributions may be aggregated into a joint distribution to increase data volume and use more complex models \citep{zhuang2020comprehensive}.

An analysis of non-IID machine learning methods that considers the unique characteristics of HRI datasets and how they interact with each method's data handling procedure is presented. Relevant methods within each category are briefly discussed to assess their merit for estimating human workload for unknown tasks using the developed criteria (i.e., portability, model complexity, and adaptability).

% Many of these methods have been successfully applied to task recognition using wearable sensors (e.g., \citep{faridee2022strangan, li2021supervised, liu2021few, soleimani2021cross}), which shares core commonalities with workload estimation. Thus, these methods provide insight into the capabilities of the underlying machine learning models to account for distribution shifts in relevant application domains. Prior work has primarily focused on evaluating standard IID methods' ability to generalize to new situations, where many methods fail to generalize to new tasks (e.g., \citep{appel2021cross, boring2020continuous, ke2015towards, zhang2018learning}). Relying on standard IID methods ignores the distribution shifts of the physiological signals and workload components as tasks change. No prior work modeled workload estimation for unknown tasks as a non-IID problem. An appropriate non-IID machine learning method for the workload estimation for unknown tasks must be identified, so that the resulting workload estimation algorithm can account for distribution shifts across tasks.

\subsection{Domain Adaptation}

% \begin{figure*}
%     \centering
%     \includegraphics[scale=0.28]{figs/domain_adaptation.png}
%     \caption{Examples images from the Bike and Laptop categories in Amazon, DSLR, Webcam, and Caltech-256 databases. Adapted from Figure 1 of \citep{wang2018deep}.}
%     \label{fig:da}
% \end{figure*}

Domain adaptation considers scenarios where training and testing data contain the same features (e.g., physiological signals), but are sampled from different distributions. 

%\vspace{2.5pt}

\begin{definition*}\textbf{(Domain Adaptation)} Given a source domain \(D_S\) and a learning task \(L_S\), as well as a target domain \(D_T\) and learning task \(L_T\), \emph{domain adaptation} aims to improve the target predictive function \(f_T(x)\), where \(D_S \neq D_T\) \citep{singhal2023domain}. 
\end{definition*}

%\vspace{2.5pt}
This distribution shift type is called a domain shift, or a covariate shift. An intuitive example is a machine learning-based workload estimation method trained on data collected in an climate controlled environment (i.e., source domain) and tested on data in an outdoor environment (i.e., target domain). Extreme temperature differences will have an impact on an individual's workload, but also alter their physiological responses to nominal workload conditions \citep{hancock2003effects}.
% For example, consider multiple datasets with images of the same objects \citep{wang2018deep}. Changes in the images' lighting, background, or orientation represent meaningful difference in the input features and are prototypical examples of a domain shift. 
Domain adaptation is a non-IID machine learning method that seeks to minimize performance degradation in the presence of domain shift. It is important to note that domain adaptation techniques focus on distribution shifts between one source and one target domain (i.e., two distributions), as shown in Figure \ref{fig:dist_shift}. 

Domain adaptation has also been called \emph{transductive transfer learning}, as many techniques require the presence of source and target domain data during training to perform well \citep{pan2009survey}. Further, there are two forms of domain adaptation: supervised \citep{wang2018deep} and unsupervised \citep{wilson2020survey}. Supervised domain adaptation requires that labels be available for \(L_T\), while there are no labels available for \(L_T\) for unsupervised domain adaptation. There tends to be significantly more data available in source domains, and using labeled data from both domains typically results in overfitting to the source distribution \citep{wang2018deep}. Thus, the majority of prior work has focused on unsupervised domain adaptation. 

Domain adaptation has been applied in broad range of HRI application domains, including task recognition \citep[e.g.,][]{alajaji2023domain, an2021adaptnet, faridee2022strangan, zakia2021force, li2022mass}, workload estimation \citep[e.g.,][]{zhou2022cross, zhou2023cross, albuquerque2022estimating}, and emotion recognition \citep[e.g.,][]{li2022dynamic, latif2022self, guo2023multi, he2022adversarial}. The vast majority of these applications use adversarial learning techniques \citep{li2021supervised, zhou2023cross, he2022adversarial}, though self-supervised and feature representation learning strategies have also been moderately successful \citep{li2022dynamic}. 

% Adversarial domain adaptation consists of three model types: (1) feature extractor, (2) domain discriminator, and (3) label predictor. Seminal work integrated adversarial learning and domain adaptation as a mini-max game \citep{ganin2016domain}. This work demonstrated that domain adaptation can be achieved by jointly optimizing a feature extraction model to maximize the difference of latent features from source and target domain data, via a domain classifier, and to minimize the error in predicted values for learning tasks, via a label predictor. This joint optimization enables the effective learning of a useful latent feature representation. Prior work demonstrated that replacing the discriminative feature extraction models with generative feature extraction models is also an effective means of domain adaptation (e.g., \citep{balaji2020robust, liu2016coupled, li2020model}). 

Most domain adaptation techniques exhibit \emph{one-to-one portability} \citep{balaji2020robust, liu2016coupled, li2020model}, but multi-target domain adaptation (i.e., one-to-many portability) is an active research area \citep{gholami2020unsupervised}. Domain adaptation techniques primarily rely on deep neural networks \citep{zhou2022cross, zhou2023cross, li2022dynamic} consisting of millions of parameters; thus, exhibiting a \emph{high model complexity}. Domain adaptation techniques exhibit varying adaptability, where some approaches require one example in the target domain \citep{yue2021prototypical}, while others require thousands \citep{li2021supervised}. Further, access to source and target data during training is a core requirement for all domain adaptation techniques, making them impractical for real-world HRI. \emph{One-to-one portability, high model complexity, and low adaptability} are undesirable properties; thus, domain adaptation techniques are not viable for workload estimation of unknown tasks.

\subsection{Transfer Learning} 

Transfer learning aims to improve the performance of a machine learning model for a target domain by transferring knowledge from a different, but related source domain.

%\vspace{2.5pt}

\begin{definition*}(\textbf{Transfer Learning}) Given a source domain \(D_S\), and a learning task \(L_S\), a target domain \(D_T\), and learning task \(L_T\), \emph{transfer learning} aims to improve the learning of a target predictive function \(f_T(x)\) in \(D_T\), using the knowledge in \(D_S\) and \(L_S\), where \(D_S \neq D_T\) and \(L_S \neq L_T\) \citep{zhuang2020comprehensive}.
\end{definition*}

%\vspace{2.5pt}

Transfer learning is characterized by a label shift \citep{huh2016makes}, either in the relative distribution of output values or a complete change in the available labels. Consider the field of affective computing's use of physiological signals \citep{egger2019emotion}. Transfer learning techniques attempt to transfer knowledge from a model trained to recognize positive emotions (e.g., awe) to a model attempting to recognize negative emotions (e.g., anger, disgust). These techniques are widely applied within HRI application domains (e.g., human task recognition \citep{ding2018empirical, li2021toward}, emotion recognition \citep{li2019multisource, quan2023eeg, ma2023cross, nguyen2023meta}).

% Transfer learning techniques are capable of transferring knowledge from a model trained to identify everyday objects (e.g., cats, planes) in images to a model performing medical image analysis (e.g., ultrasounds, x-rays) \citep{morid2021scoping}. Similar to domain adaptation, transfer learning has been widely applied to HRI (e.g., human task recognition \citep{ding2018empirical, li2021toward}, emotion recognition \citep{li2019multisource, quan2023eeg, ma2023cross, nguyen2023meta}). 

There are many transfer learning techniques that vary based on how knowledge is transferred: instance-based \citep[e.g.,][]{chen2022feature, wang2019minimax, wang2019instance}, feature-based \citep[e.g.,][]{ren2019new, stuber2018feature, wu2021research}, and parameter-based \citep[e.g.,][]{fu2021personalized, huh2016makes, soleimani2021cross, link2022wearable}.  Instanced and feature-based techniques have had moderate levels of success, but are seldom used in modern applications when compared to parameter-based techniques \citep{weiss2016survey}.

Parameter-based methods have two phases: pre-training and fine-tuning. \emph{Pre-training} optimizes a machine learning model on data from the source domain, \(D_S\), to solve the source learning task \(L_S\). \emph{Fine-tuning} trains the same model on data from the target domain, \(D_T\), but only updates a subset of the parameters. This two-step training procedure enables the network to learn a useful latent representation during pre-training, and a task-specific transformation of that representation during fine-tuning. This approach has been successful in computer vision \citep{huh2016makes}, robotics \citep{lee2021causal}, and task recognition \citep[e.g.,][]{an2023transfer, ray2023transfer, pavliuk2023transfer}.

% and human activity recognition (e.g., \citep{cook2013transfer, fu2021personalized, link2022wearable, soleimani2021cross}).

Many of these methods use deep learning architectures to transfer knowledge across source/target domain pairs, requiring thousands of examples to be successful. Further, there is evidence to suggest that the \emph{model complexity} is inversely related to both \emph{portability} and \emph{adaptability} \citep{zhuang2020comprehensive, brown2020language}. Zhuang et al. evaluated a range of deep learning architectures on several tasks, all of which exhibited \emph{one-to-one portability, high model complexity, and low adaptability} \citep{zhuang2020comprehensive}. Similar techniques were developed to solve a wide range of natural language problems based on only a few examples \citep{brown2020language}. This \emph{high adaptability} is in direct contrast to the other transfer learning techniques, where the primary difference is the \emph{model complexity}. A transformer deep learning architecture was trained to predict the next word in a sentence \citep{brown2020language}. This network was the foundation of several downstream networks that performed tasks, including machine translation, question answering, and arithmetic word problem-solving. The architecture consisted of 175 billion parameters, requiring significantly more data to train than the parameter-based techniques that use smaller networks (i.e., tens of billions of data points). These results suggest that a higher \emph{model complexity} can directly improve \emph{portability} and \emph{adaptability}.

A core limitation of workload estimation is data volume, as gathering large volumes of wearable sensor data is logistically impractical; thus, \emph{model complexity} is constrained. It is also desirable that workload estimation algorithms exhibit \emph{many-to-many portability}, as models may need information across multiple individuals and tasks. Thus, transfer learning is not a viable option.

\subsection{Test-time Adaptation}

Domain adaptation techniques learn to anticipate domain shifts by simultaneously training on \(D_S\) and \(D_T\); however, target domain data may not be available during training. Test-time adaptation techniques seek to overcome this limitation.

%\vspace{2.5pt}

\begin{definition*}\textbf{(Test-time Adaptation)} Given a source domain \(D_S\) and a learning task \(L_S\), as well as a target domain \(D_T\) and learning task \(L_T\), \emph{test-time adaptation} aims to improve the learning of a target predictive function \(f_T(x)\) based on an existing predictive function \(f_S(x)\), where \(D_S \neq D_T\) \citep{sun2020test}. 
\end{definition*}

%\vspace{2.5pt}

Test-time adaptation techniques seek to overcome domain shifts by separating the training process into two distinct phases: (1) training on \(D_S\) data, and (2) updating that model with \(D_T\) data. This separation enables the resulting model to adapt at test-time by allowing for the continuous updating of model parameters, as shown in Figure \ref{fig:dist_shift} \citep{sun2020test}.

%Task recognition
Prior work applied test-time adaptation techniques to egocentric task recognition \citep{plananamente2022test} and adaptive policy optimization for assistive robots \citep{he2023learning}. However, test-time adaptation techniques have not been widely applied to HRI application domains. Nevertheless, these techniques hold promise because they overcome the limitation of training on source and target domain data simultaneously. These techniques can be categorized into as either self-supervised, or entropy-based. Enumerating the algorithmic details for each of these techniques will highlight how they may be used in future human workload estimation algorithms.

%NOTE: DO NOT DELETE. THIS TEXT CONTAINS OLDER CITATIONS
% Self-supervised techniques use a dual-output neural network and a two-phase training procedure \citep{sun2020test}. Labeled data from the \(D_S\) undergoes some deterministic transformation (e.g., image rotation). The neural network is trained on this transformed data to predict both the original label (e.g., image class) and the corresponding transformation (e.g., rotation angle). The unlabelled data from \(D_T\) undergoes the same deterministic transformation. The shared feature extraction model's parameters are updated based on the output of the self-supervised proxy learning tasks. This two-phase procedure decouples training from adaptation and enables the model to adapt at test-time. Many techniques take inspiration from this work (e.g., \citep{azimi2022self, chi2021test, he2021autoencoder}); however, the development of an application domain relevant proxy task is non-trivial, limiting the generalizability of self-supervised techniques. 

% Entropy-based techniques eliminate the need for a self-supervised proxy task by using a batch processing adaptation phase (e.g., \citep{liu2021ttt, zhang2021memo, Zhou2021TrainingOT}). Batches of data from \(D_T\) are passed through the shared feature extractor model, and the Shannon Entropy \citep{shannon1948mathematical} is calculated for each batch. The adaptation phase updates the shared feature extractor to minimize the average entropy over all batches \citep{wang2020tent}.  

Self-supervised techniques use a dual-output neural network and a two-phase training procedure \citep{sun2020test}. Labeled data from the \(D_S\) undergoes some deterministic transformation (e.g., image rotation). The neural network is trained on this transformed data to predict both the original label (e.g., image class) and the corresponding transformation (e.g., rotation angle). The unlabelled data from \(D_T\) undergoes the same deterministic transformation. The shared feature extraction model's parameters are updated based on the output of the self-supervised proxy learning tasks. This two-phase procedure decouples training from adaptation and enables the model to adapt at test-time. Many techniques take inspiration from this work \citep[e.g.,][]{azimi2022self, chen2023improved, segu2023darth}; however, the development of an application domain relevant proxy task is non-trivial, limiting the generalizability of self-supervised techniques. 

Entropy-based techniques eliminate the need for a self-supervised proxy task by using a batch processing adaptation phase \citep[e.g.,][]{liu2021ttt, zhang2021memo, zhang2023domainadaptor}. Batches of data from \(D_T\) are passed through the shared feature extractor model, and the Shannon Entropy \citep{shannon1948mathematical} is calculated for each batch. The adaptation phase updates the shared feature extractor to minimize the mean entropy over all batches \citep{wang2020tent}. Other techniques use the same approach, but minimize Shannon Entropy from \(D_S\) data and \(D_T\) data \citep{jung2023cafa, seto2024realm}. Prior work demonstrated that minimizing entropy in this way allows the feature extraction model to learn domain-invariant features to improve generalization to new learning tasks \citep{adachi2023covariance}. Recent work has extended entropy-based techniques to incorporate varying methods for comparing data points between \(D_S\) and \(D_T\) \citep[e.g.,][]{song2023ecotta, niu2022efficient, zhang2023adanpc}.

Fundamentally, test-time adaptation is domain adaptation without access to target domain data during training. Therefore, test-time adaptation exhibits \emph{one-to-one portability, high model complexity, and low adaptability}. It is difficult to definitively state whether these techniques are applicable to workload estimation for unknown tasks, as the relationship between \emph{model complexity} and \emph{adaptability} for these techniques is unknown. Prior work has primarily focused on deep learning architectures, but these techniques are general enough to apply to shallow neural networks. It is possible that smaller networks may possess sufficient adaptability to account for real-world variability, but the efficacy of test-time adaptation on these networks has not been evaluated.
 
\subsection{Continual Learning}

Continual Learning techniques train a model to perform a sequence of non-IID learning tasks, where each task is characterized by a unique domain shift (see Figure \ref{fig:dist_shift}). 

%\vspace{2.5pt}

\begin{definition*}\textbf{(Continual learning)} Given a sequence of domains \(D\) = \(D_{s_1}, \dots, D_{s_n}\), where \(n>0\), and a sequence of learning tasks \(T\) = \(L_{s_1}, \dots, L_{s_n}\), where learning task \(L_{s_i}\) corresponds to domain \(D_{s_i}\), \emph{continual learning} aims to optimize the predictive function \(f_i(x)\) using knowledge from previously seen domains \(D_j \in D\), where \(i \leq j\) \citep{chen2018lifelong}.
\end{definition*} 

%\vspace{2.5pt}

%NOTE: DO NOT DELETE THIS TEXT CONTAINS OLDER CITATIONS
% These techniques gradually acquire knowledge to optimize performance on the latest learning task. Continual learning is also referred to as lifelong learning, sequential learning, or online learning \citep{mai2022online}. There is a wide range of continual learning techniques, including knowledge distillation (e.g., \citep{chen2022incremental, wu2019large, zhu2021prototype}), memory-based methods (e.g., \citep{cai2021online, derakhshani2021kernel}), and parameter isolation (e.g., \citep{lomonaco2021avalanche, serra2018overcoming}).

These techniques gradually acquire knowledge to optimize performance on the latest learning task, and some techniques add the constraint of avoiding performance degradation on all prior tasks \citep{smith2023closer}. Continual learning is also referred to as lifelong learning, sequential learning, or online learning \citep{mai2022online}. There is a wide range of techniques, including knowledge distillation \citep[e.g.,][]{chen2022incremental, wu2019large, zhu2021prototype}, memory-based methods \citep[e.g.,][]{cai2021online, derakhshani2021kernel}, and parameter isolation \citep[e.g.,][]{lomonaco2021avalanche, serra2018overcoming}.

Continual learning techniques are one of the more popular techniques within HRI, having been successfully applied to problems of human task recognition using wearable sensors \citep[e.g.,][]{hasan2015continuous, ye2019lifelong, ashry2020charm, leite2022resource}, affective computing \citep[e.g.,][]{churamani2020continual, gao2023ssa, churamani2022continual}), and adaptive HRI \citep[e.g.,][]{churamani2020continual, lesort2020continual, spaulding2021lifelong}. Further, probabilistic continual learning techniques are common in the cognitive science and psychology communities to construct human decision making models \citep[e.g.,][]{austerweil2013nonparametric, collins2013cognitive, liew2016appropriacy, kanwal2022towards, navarro2006modeling, niv2019learning}. Specifically, Dirichlet process mixture models that function as a non-parametric Bayesian clustering algorithm that can recognize existing classes and detect out-of-distribution classes \citep{li2019tutorial}. Recent work demonstrated that these techniques can be combined with deep-learning models to improve performance for computer vision tasks \citep[e.g.,][]{jerfel2019reconciling, xu2024context, willes2022bayesian}.

Conventional continual learning assumes the new learning tasks arrive one at a time, and the data in the domains is stationary \citep{de2021continual}. Thus, continual learning techniques can be trained offline, as the number of tasks is known a priori. These techniques are called multi-head techniques, as separate models are trained for each learning task \citep{chaudhry2018riemannian}. However, knowing all the tasks a priori is not always realistic. Other continual learning techniques make no assumptions about the distribution of learning tasks, and seek to identify the task online \citep{chaudhry2018riemannian}. These single-head techniques train a single model is trained for all tasks.

Learning tasks within a sequence can either be independent or dependent. Independent learning tasks are separate and may not share similarities. Dependent learning tasks represent a sub-sequence of learning tasks that must be executed in a particular order, where learning task \(L_i\) holds key information about learning task \(L_{i+1}\). Many of these techniques rely heavily on deep neural networks, similar to both domain adaptation and transfer learning techniques \citep{mundt2023wholistic, ao2023continual}.

The \emph{model complexity} of continual learning techniques varies; however, there is evidence that \emph{model complexity} is not directly tied to performance, as the machine learning model choice is problem specific (e.g., Gaussian processes \citep{spaulding2021lifelong}, shallow neural networks \citep{derakhshani2021kernel}, and deep networks \citep{mai2022online}). Many techniques exhibit \emph{low adaptability}, requiring thousands of examples to perform well \citep{mai2022online}. \emph{Portability} is challenging to define for continual learning, as it exhibits \emph{many-to-many portability}, but does so only in a sequential fashion. \emph{Many-to-many portability} is desirable, but the sequential nature is an unnecessary constraint. Additionally, some continual learning methods only optimize performance on the latest learning tasks, while other optimize performance on all prior learning tasks. \emph{Portability} and \emph{adaptability} are key qualities required to estimating workload successfully for unknown tasks; thus, continual learning is a subpar option.

\subsection{Domain Generalization}

Domain generalization considers the setting where training data from multiple source domains is characterized by the same features, but sampled from different distributions \citep{li2018learning}. Specifically, domain generalization is the process of transferring knowledge from several related source domains and applying it to previously unknown domains \citep{muandet2013domain}, shown in Figure \ref{fig:dist_shift}. 

%\vspace{2.5pt}

\begin{definition*}\textbf{(Domain Generalization)} Given a set of source domains \(DS\) = \(D_{s_1}, \dots, D_{s_n}\) where \(n > 0\), a target domain, \(D_t\), a set of source tasks \(LS\) = \(L_{s_1}, \dots, L_{s_n}\) where \(L_{s_i} \in LS\) corresponds with \(D_{s_i} \in DS\), and a target task \(L_t\) that corresponds to \(D_t\), \emph{domain generalization} aims to optimizes a target predictive function \(f_t(x)\) in \(D_t\), where \(D_t \notin DS\) \citep{li2018learning}.
\end{definition*}  

%\vspace{2.5pt}
The goal is to leverage unique cross-domain information to improve the machine learning model's generalization. These methods assume that constructing a joint distribution of source domains aggregates away key information, and that the varying quality and relevance of data across source domains helps inform a model in unknown target domains. A key feature of domain generalization is the lack of labeled target domain samples. This setting exhibits natural similarities to human workload estimation for unknown tasks, as novel tasks will incur a novel physiological response not encountered in the training dataset (i.e., domain shift). Additionally, it is unlikely that labeled data will be available when attempting to estimate workload for unknown tasks in real-world environments. Domain generalization techniques can be broadly classified into three categories: domain-invariant representation learning \citep[e.g.,][]{suh2023tasked, li2022new, gu2023generalizable}, data manipulation \citep[e.g.,][]{ilse2020diva, li2021simple, yue2019domain}, and learning strategy \citep[e.g.,][]{khandelwal2020domain, sicilia2023domain, liu2020shape}.

%NOTE: DO NOT DELETE OLD TEXT THAT CONTAINS OLDER CITATIONS
% Domain generalization techniques can be broadly classified into three categories: domain-invariant representation learning (e.g., \citep{hu2020domain, li2022new, zhao2020domain}), data manipulation (e.g., \citep{ilse2020diva, li2021simple, yue2019domain}), and learning strategy (e.g., \citep{khandelwal2020domain, li2018learning, liu2020shape}).

The primary technique for domain-invariant representation learning is domain alignment, which seeks to minimize the difference across source domains, such that a latent representation consisting of only essential features is learned \citep{hu2020domain}. Domain alignment techniques measure this difference in many ways, including minimizing moments \citep{ghifary2016scatter}, contrastive loss \citep{kim2021selfreg}, or maximum mean discrepancy \citep{li2018domain}. A key difference between domain-adversarial techniques for domain generalization and domain adaptation is that the discriminator must differentiate between multiple source domains. 

Data manipulation techniques seek to artificially enhance the size and diversity of a dataset via data augmentation or generation. Data augmentation performs deterministic transformations over existing data to create new examples \citep{otalora2019staining}. However, data augmentation is not domain-agnostic, as useful transformations are specific to an application domain. Data generation uses generative machine learning models (e.g., Variational Autoencoders \citep{ilse2020diva}, Generative Adversarial Networks \citep{li2021simple}) to create completely new training examples. These techniques are domain-agnostic, but the underlying models are hard to train and require a high data volumes \citep{goodfellow2020generative}. These techniques cannot be applied to workload estimation, as generating realistic physiological metrics that correctly correspond to the workload values is non-trivial. 

Learning strategy-based domain generalization techniques seek to improve generalization through novel training methodologies, including: 1) Self-supervised learning 2) Ensemble learning, and 3) Meta-learning. Self-supervised learning techniques for domain generalization do not differ substantially from self-supervised techniques for test-time adaptation. Ensemble learning is the process of training multiple machine learning models with varying weight initializations or training data splits, then aggregating the models' outputs to make predictions. These techniques are notoriously computationally expensive \citep{zhou2012ensemble} and are not viable solutions for real-time HRI domains.

\begin{figure*}
    \centering
    \includegraphics[width=\textwidth,height=6cm]{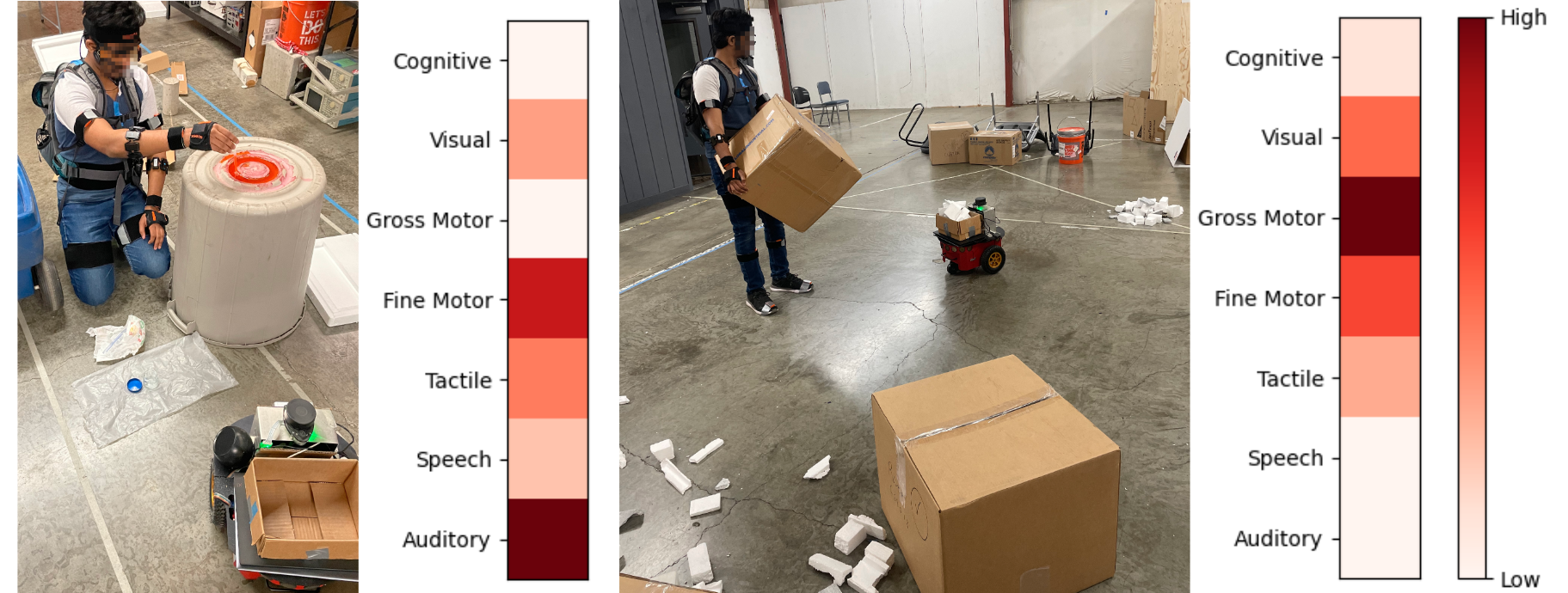}
    \caption{A human-robot team conducting two tasks, where each task is characterized by a unique distribution across the seven workload components. These tasks were implemented as a part of an experimental design for developing a multi-dimensional workload estimation method for unknown tasks \citep{bs2024design}.}
    \label{fig:dg_fsl}
\end{figure*}

Meta-learning trains a machine learning model on data across source domains, such that that model can be efficiently updated to operate in arbitrary target domains \citep{finn2017model}. Broadly, meta-learning is `learning to learn' \citep{hospedales2020meta}. Domain generalization typically leverages parameter-based meta-learning techniques (e.g., Model Agnostic Meta-Learning (MAML) \citep{finn2017model}) that seeks to construct a neural network that can be adapted based on a few examples (see Figure \ref{fig:dg_fsl}). 

A meta-model, parameterized by \(\theta\), is trained to predict an initialization for the parameters of another model, \(\phi_i\), that performs the learning task \(L_i\). Both \(\theta\) and \(\phi_i\) typically represent a neural network's weights, but can also represent a subset of the network's weights, or the another model's parameters. Each source domain \(D_i\) and learning task \(L_i\) is deconstructed into a collection of mini-datasets. These mini-datasets are defined by the support set, \(S_i\), and the query set, \(Q_i\), which are small training and test sets, respectively. The support set consists of a few (e.g., 1, 5, 10) examples that the meta-model uses to predict the labels for the query set. The task-specific model's error is twice differentiated and backpropagated through both models to update \(\theta\). This problem formulation constructs the learning process, such that the model learns how to adapt its parameters based on a few examples. This technique is popular in meta-learning \citep{hospedales2020meta}, and is the foundation for many techniques \citep[e.g.,][]{antoniou2018train, bertinetto2018meta, finn2018probabilistic, nichol2018first, yoon2018bayesian}. An alternative approach learns the parameters of an optimizer. An LSTM was trained to learn the exact optimization process of a neural network, such that the new network converged using only a few samples \citep{ravi2016optimization}. The technique Meta-Stochastic Gradient Descent combined the LSTM approach and MAML to learn an initialization of the parameters, and an optimal optimizer \citep{li2017meta}.

The majority of domain generalization techniques exhibit \emph{many-to-many portability} by construction. Data manipulation is characterized by \emph{high model complexity}, due to its reliance on generative deep learning architectures \citep{zhou2020learning}. Additionally, these techniques exhibit a range of \emph{adaptability}, where some instances require several thousands of examples for adaptation \citep{xu2021generative}, and others only require a few examples \citep{chen2022discriminative}. MAML and other meta-learning techniques exhibit \emph{high adaptability} and are general enough to apply to shallow neural networks; thus, exhibiting a \emph{low model complexity}. Therefore, meta-learning based domain generalization techniques may be applicable to workload estimation for unknown tasks; however, these techniques have been shown to be computationally expensive at both training and inference times \citep{antoniou2018train}.

%TODO: Refine this caption a bit more.

\subsection{Few-shot Learning}

Few-shot learning aims to train a machine learning model using data across multiple source domains, such that the model can be efficiently adapted to make predictions in some unknown target domain \citep{wang2020generalizing}. 

%\vspace{2.5pt}

\begin{definition*}\textbf{(Few-shot Learning)} Given a set of source domains \(DS\) = \(D_{s_1}, \dots, D_{s_n}\) where \(n > 0\), a target domain, \(D_t\), a set of source learning tasks \(LS\) = \(L_{s_1}, \dots, L_{s_n}\) where \(L_{s_i} \in LS\) corresponds with \(D_{s_i} \in DS\), and a target learning task \(L_t\) that corresponds to \(D_t\), \emph{few-shot learning} optimizes the target predictive function  \(f_t(x)\) in \(D_t\), where \(D_t \notin DS\) and \(L_t \notin LS\) \citep{wang2020generalizing}.
\end{definition*}  

%\vspace{2.5pt}

Few-shot learning is characterized by both a shift in the input distribution (i.e., domain shift) and a shift in the target domain labels (i.e., label shift), as indicated in Figure \ref{fig:dist_shift}. This problem formulation is similar to domain generalization, but with additional complexity. A domain shift results from the novel physiological response that an unknown task can induce, but a label shift is a new workload value. An illustrative example demonstrating the primary difference is provided in Figure \ref{fig:dg_fsl}. The left task has high auditory and fine motor workload, but a low gross motor workload. The right task has a high gross motor workload, but a low auditory workload. Training a model to estimate gross motor workload on the left task will result in a model that likely performs poorly when estimating workload for the right task, as the range of gross motor workload values is too different. Few-shot learning techniques overcome the additional complexity of handling multiple distribution shifts simultaneously by requiring labeled target domain data.

Prior work primarily focused on applying meta-learning techniques to solve few-shot learning problems. These techniques overlap significantly with meta-learning for domain generalization. Meta-learning constitutes the vast majority of popular few-shot learning algorithms (e.g., MAML \citep{finn2017model}, Prototypical network \citep{snell2017prototypical}, Conditional Neural Processes \citep{garnelo2018conditional})). Meta-learning techniques broadly fall into three categories: 1) parameter-based, 2) metric-based, and 3) probabilistic. Parameter-based meta-learning algorithms were discussed in the Domain Generalization section. 

Metric-based meta-learning attempts to learn a robust latent representation across source domains, such that data points can be directly compared \citep{vinyals2016matching, zhou2023attribute, li2023deep}. Two neural networks are trained to encode data, from the source and target domain, into a latent feature space. One neural network, \(g_{\theta}\), encodes the support set (i.e., source domain data), and a second, \(f_{\theta}\), encodes the query set (i.e., target domain data). Typically, multiple support set data points are passed through the networks simultaneously to enable the calculation of a range of complex distance metrics.

% \begin{figure*}
%     \centering
%     \includegraphics[scale=0.35]{figs/neural_processes.png}
%     \caption{A comparison of two basic neural process types: (a) conditional neural processes and (b) latent neural process. Green boxes represent input and output values, blue boxes represent a neural network, yellow boxes represent aggregated latent features, and the pink circle represent an aggregation function. This figure is adapted from Figure 3 in \citep{jha2022neural}.}
%     \label{fig:neural_process}
% \end{figure*}

% \begin{figure*}
% \begin{minipage}[b]{0.5\textwidth}
%         \centering
%         \includegraphics[height=3.5cm, width=0.8\textwidth]{figs/CNP.png}\\
%         \subcaption{Conditional Neural Process}
%     \end{minipage}%
%     \hfill
%     \hfill
%     \begin{minipage}[b]{0.5\textwidth}
%         \centering
%         \includegraphics[height=3.5cm, width=0.9\textwidth]{figs/LNP.png}\\
%         \subcaption{Latent Neural Process}
%     \end{minipage}%
%      \caption{A comparison of two basic neural process types: (a) conditional neural processes and (b) latent neural process. Green boxes represent input and output values, blue boxes represent a neural network, yellow boxes represent aggregated latent features, and the pink circle represent an aggregation function. This figure is adapted from Figure 3 in \citep{jha2022neural}.}
%      \label{fig:neural_process}
% \end{figure*}

Matching networks perform a pairwise comparison between latent features using cosine similarity \citep{vinyals2016matching}. Prototypical networks extend this technique by using additional data points to calculate the centroid of each class' location within the latent space \citep{snell2017prototypical}. Each class is represented by \(N\) examples, and when \(N\) is equal to one, prototypical networks are identical to matching networks \citep{snell2017prototypical}. Relation networks decompose the problem into separate modules: an embedding module and a relation module \citep{sung2018learning}. The embedding module is a neural network designed for feature extraction and the relation module is a non-linear classifier trained on the latent feature representation. Adaptive subspace networks use singular-value decomposition on the latent features to construct the basis vectors of subspaces for each class \citep{simon2020adaptive}. This technique maximizes the distance between subspaces by using a Grassmannian geometry-based objective function. Recent work modeled encoded data points probabilistically to improve information contained within latent features \citep[e.g.,][]{kim2019variational, schonfeld2019generalized, zhang2019variational}. 

% Recent work modeled encoded data points probabilistically to improve the information contained within latent features (e.g, \citep{kim2019variational, schonfeld2019generalized, yang2016autoencoder, zhang2019variational}). 

% \textcolor{red}{TODO: Make this section shorter}
Generally, probabilistic meta-learning typically can be viewed as the combination of a stochastic process (i.e., Gaussian Process \citep{rasmussen2003gaussian}) and a neural network. Combining these two different machine learning models takes the best of both worlds by leverage the Gaussian processes' Bayesian framework and the neural network's ability to extract latent features from large datasets \citep{jha2022neural}. Probabilistic meta-learning techniques fall into two broad categories: Neural Processes \citep{garnelo2018neural}, and Deep Kernel Learning \citep{wilson2016deep}.

Neural processes pass the data through a neural network to learn a latent representation. Each support set data point is passed through an aggregation function to calculate the latent feature. This aggregation is performed to maintain indifference under the permutation of latent features, which helps maintain a key theoretical property (i.e., the consistency condition) of stochastic processes \citep{jha2022neural}). Another neural network uses that latent representation and one or more data points (e.g., one-shot vs. few-shot) from the target domain to learn the parameters of a Gaussian Process. There are two main neural process types: latent neural processes \citep{garnelo2018neural} and conditional neural processes \citep{garnelo2018conditional}. Conditional neural processes have received the most attention, where many techniques incorporate different neural network architectural components with varying degrees of success \citep[e.g.,][]{gu2023general_neural_fields, holderrieth2021equivariant, bruinsma2023autoregressive, kim2022neural, wang2022np, ye2022contrastive, dexheimer2023learning, pandey2023evidential}).

Deep kernel learning techniques take an alternative approach by constructing an end-to-end model that simultaneously trains a neural network and a Gaussian process, training the network to learn a custom kernel function for that Gaussian process \citep{wilson2016deep}. Deep kernel transfer first applied this technique to few-shot learning by maximizing the marginal likelihood across multiple tasks \citep{patacchiola2020bayesian}. This technique learns over a large number of small, but related, examples that train the machine learning model to approximate a prior distribution to transfer knowledge between learning tasks. Deep kernel transfer quantifies uncertainty by maintaining a distribution over parameters of the neural network, which can become computationally expensive for larger networks. 
% P\'{o}lya-Gamma augmentation and softmax approximation were explored to overcome this limitation by maintaining a distribution over functions \citep{snell2021bayesian}. 
% Deep mean function learning has also been applied as a meta-learning technique, where a neural network is used to learn a custom mean function for a Gaussian process \citep{rothfuss2021pacoh}. 
Similar techniques have received increased attention, as neural networks have been combined with a range of probabilistic models \citep[e.g.,][]{rothfuss2023meta, wei2023adaptive, brankestad2023variational, maraval2024end, iwata2024meta, lee2023efficient}. However, recent results demonstrate that many deep kernel learning algorithms tend to produce unreliable uncertainty estimates in some application domains \citep{van2021feature}. Incorporating modern deep kernel learning techniques and practices has the potential to improve the generalization of the presented meta-learning techniques \citep{liu2022simple}.

% Neural Processes and deep kernel learning-based techniques all exhibit many-to-many portability and high adaptability by construction. Further, prior work demonstrated that both neural processes and deep kernel learning-based techniques are general enough to apply to both deep-learning architectures, as well as shallow neural networks \citep{patacchiola2020bayesian} \citep{garnelo2018conditional}. The additional complexity of the Gaussian Process is negligible in comparison to that of the neural network, as Gaussian processes are non-parametric models \citep{rasmussen2003gaussian}. Thus, model complexity is problem-dependent, similar to metric-based techniques. 

% %\vspace{2.5pt} \break
Few-shot learning techniques exhibit \emph{many-to-many portability and high adaptability}. Most few-shot learning techniques were developed using computer vision benchmarks \citep[e.g.,][]{garcia2018few, snell2017prototypical, vinyals2016matching}, which requires deep learning architectures. However, prior work demonstrated that few-shot learning techniques' success is directly tied to a good feature extraction model \citep{tian2020rethinking}. Therefore, it stands to reason that \emph{model complexity} is directly related to the learning task's complexity. Meta-learning techniques have been proven to be successful solutions to both few-shot learning and domain generalization. Further, these techniques exhibit desirable properties for all three presented criteria. Therefore, meta-learning presents a promising option for develop workload estimation for unknown tasks solutions and merit future investigation.

% Refine this section to highlight the data-specific considerations to understand how we can more appropriately model these complex data charcteristics.
\section{Discussion}

Workload estimation for unknown tasks relies on noisy and varied metrics, as different physiological signals and workload components are particularly important for each task. Understanding the distribution across workload components is critical to adapting robot behavior to maximize the human-robot team's performance.

Quantifying uncertainty helps enumerate this distribution by contextualizing component-specific workload estimates, as components may rely on overlapping metrics (e.g., fine-motor and tactile). Viewing candidate machine learning methods through the three developed criteria (i.e., portability, model complexity, and adaptability) facilitates discussing for how each candidate addresses the core concerns of estimating human workload for unknown tasks (see Evaluation Criteria section).

Domain adaptation and transfer learning are popular methods, but are most applicable to problems where knowledge can be transferred between data-rich source and data-sparse target domains, and where all data is gathered prior to training (i.e., \emph{one-to-one portability} and \emph{low adaptability}). These methods are not well suited for estimating a human's workload for unknown tasks, as it is impractical to gather a significant amount of data for new tasks (i.e., target domain) prior to training the model. Additionally, these techniques do not adapt machine learning models in real time. Test-time adaptation overcomes this particular constraint, but exhibits \emph{one-to-one portability} and \emph{low adaptability}. Existing test-time adaptation techniques applied to smaller neural networks may exhibit higher adaptability, but further investigation is required. Regardless, none of these techniques are well suited for estimating human workload in the presence of unknown tasks.

Continual learning has been frequently proposed as a potential solution to non-IID machine learning problems within HRI, but are not applicable to workload estimation for unknown tasks due to their \emph{poor portability}. Incorporating information for an indeterminant number of tasks, while avoiding performance degradation on all prior tasks are additional constraints that likely make the machine learning problem unnecessarily difficult. These techniques learn how to transform the model between task pairs, whereas other techniques (e.g., domain generalization, few-shot learning) learn to transform from a common prior. Consider an application domain with \(N\) tasks. Continual learning must learn \(N!\) task-pair independent updates, as the tasks' ordering influences the learning process. Domain generalization and few-shot learning techniques only learn \(N\) updates, as they can trace each task to a single common prior. These additional portability-related constraints make continual learning a subpar option.

Domain generalization and few-shot learning exhibit desirable properties for all criteria: \emph{high adaptability, one-to-one portability, and low model complexity}. These techniques are designed to update a machine learning model based on limited and highly variable data, and are flexible enough to change the model complexity based on the learning task’s complexity. Multiple domain shifts from a common prior is the most directly related metaphor for task-based variability, as the contribution of each physiological signal to each workload component and the contribution of each workload component will vary between tasks. Therefore, domain generalization is the most suited for workload estimation for unknown tasks when there is no label shift. Few-shot learning is more applicable to scenarios where there are significant changes to the range of workload values between tasks (i.e., label shift). Meta-learning techniques sit at the intersection of these two categories and are general enough to apply to either learning task.

Utilizing individual workload components (i.e., cognitive, speech, auditory, visual, gross motor, fine motor, tactile) to produce measures of overall workload is fundamentally a regression problem \citep{heard2019diagnostic, bs2022visual, bs2022physical}. Only a limited number of techniques (e.g., probabilistic meta-learning) have been applied to regression problems with those applications being focused on toy problems and contrived benchmarks. Probabilistic meta-learning techniques (e.g., deep kernel transfer, conditional neural process) are particularly promising for solving workload estimation for unknown tasks, as producing continuous values for each workload component is a key requirement for informing an adaptive teaming system.

Existing workload estimation datasets frequently struggle to capture the non-IID nature of human behavior, such that probabilistic meta-learning techniques' ability to accurately estimate workload for novel tasks can be assessed. It is likely that new human subjects evaluations are required to generate the necessary data, but designing evaluations that appropriately capture the variability of human behavior is non-trivial \citep{bs2024design}. These evaluations must incorporate diverse, but ecologically valid tasks characteristic of the real world, while also considering the degree of difference between IID (i.e. known) and non-IID tasks (i.e., unknown) that these techniques can handle. Understanding how probabilistic meta-learning techniques can estimate workload accurately for similar unknown tasks, as well as dissimilar unknown tasks is crucial to understanding their real-world viability. Additionally, analyzing how the differences between tasks, human-robot interactions, and robot capabilities further informs how these techniques can be successfully deployed. Regardless, these techniques meet all three criteria and are the most promising path forward for solving workload estimation for unknown novel tasks; thus, merit further investigation.

% Stronger link between theoretical analysis and practical applications are needed.
\section{Conclusion}

Developing models that can estimate the human's workload is critical to developing dynamic human-robot teams that can operate in uncertain, dynamic environments. The variability of these environments is characterized by varying human-robot teaming dynamics, diverse tasks executions, and different operating conditions for the team itself. A robot seeking to adapt its behavior to best assist its human teammate must estimate the human's workload accurately in any of these conditions; however, real-world variability often violates the IID assumption making standard machine learning methods infeasible. This manuscript developed three criteria (i.e., portability, model complexity, and adaptability) and presented an analysis of relevant non-IID machine learning techniques. Numerous non-IID machine learning techniques exist, so this manuscript assessed how each technique's machine learning considerations interact with the unique characteristics of HRI datasets (e.g., workload estimation for unknown tasks). A deeper understanding of these interactions highlight the specific requirements for a particular application, which is useful for guiding future empirical research. This manuscript's analysis argues that domain generalization and few-shot learning techniques hold promise in developing models that can estimate a human's workload for unknown tasks.

%%Harvard (name/date)
\bibliographystyle{SageH}
%%Vancouver (numbered)
% \bibliographystyle{SageV}
\bibliography{main.bib}

\section{Copyright statement}
Please  be  aware that the use of  this \LaTeXe\ class file is
governed by the following conditions.

\subsection{Copyright}
Copyright \copyright\ \volumeyear\ SAGE Publications Ltd,
1 Oliver's Yard, 55 City Road, London, EC1Y~1SP, UK. All
rights reserved.

\begin{acks}
The graduate students have been supported by National Aeronautics and Space Administration through University Space Research Association aware 904186092. The contents are those of the authors and do not represent the official views of, nor an endorsement, by the National Aeronautics and Space Administration. 

The presented work was partially supported by ONR grants N00024-20-F-8705 and N00014-21-1-2052. The views, opinions, and findings expressed are those of the authors and are not to be interpreted as representing the official views or policies of the Department of Defense or the U.S. Government.
\end{acks}

\end{document}